\documentclass[journal]{IEEEtran}

\usepackage{ifpdf}

\usepackage{booktabs}
\usepackage{multicol}
\usepackage{booktabs}
\usepackage{cite}

\ifCLASSINFOpdf
  \usepackage[pdftex]{graphicx}

\else

  \usepackage[dvips]{graphicx}

\fi

\usepackage{amsmath}
\usepackage{amssymb} 

\usepackage{algorithmic}
\usepackage{array}

\ifCLASSOPTIONcompsoc
 \usepackage[caption=false,font=normalsize,labelfont=sf,textfont=sf]{subfig}
\else
 \usepackage[caption=false,font=footnotesize]{subfig}
\fi

\usepackage{fixltx2e}

\usepackage{stfloats}

\ifCLASSOPTIONcaptionsoff
 \usepackage[nomarkers]{endfloat}
\let\MYoriglatexcaption\caption
\renewcommand{\caption}[2][\relax]{\MYoriglatexcaption[#2]{#2}}
\fi

\usepackage{url}

\begin{document}

\title{Haptic Stiffness Perception Using Hand Exoskeletons in Tactile Robotic Telemanipulation}

\author{Gabriele Giudici$^{1}$, Claudio Coppola$^{2}$, Kaspar Althoefer$^{1}$, Ildar Farkhatdinov$^{1,3}$, Lorenzo Jamone$^{1}$
\thanks{This work involved human subjects in its research. Approval of all ethical and experimental procedures and protocols was granted by the Queen Mary Ethics of Research Committee under Application No. QMERC20.565.DSEECS24-053,
and performed in line with the Application of - Enhancing Teleoperation with Exoskeletal Gloves: Leveraging Tactile Sensing for Bilateral Haptic Stiffness Feedback.}

\thanks{$^1$ARQ (the Centre for Advanced Robotics @ Queen
Mary), School of Engineering and Materials Science, Queen Mary University of London, London, E14NS, UK (emails: \{g.giudici,k.althoefer,i.farkhatdinov,l.jamone\}@qmul.ac.uk).}%
\thanks{$^2$ Humanoid AI. email: ccop@thehumanoid.ai}
\thanks{$^3$School of Biomedical Engineering and Imaging Sciences, King's College London, London, UK.}
}

\maketitle
%
\begin{abstract}
Robotic telemanipulation—the human-guided manipulation of remote objects—plays a pivotal role in several applications, from healthcare to operations in harsh environments. 
While visual feedback from cameras can provide valuable information to the human operator, haptic feedback is essential for accessing specific object properties that are difficult to be perceived by vision, such as stiffness.
For the first time, we present a participant study demonstrating
that operators can perceive the stiffness of remote objects during real-world telemanipulation with a dexterous robotic hand, when haptic feedback is generated from tactile sensing fingertips.
Participants were tasked with squeezing soft objects by teleoperating a robotic hand, using two methods of haptic feedback: 
one based solely on the measured contact force, while the second also includes the squeezing displacement between the leader and follower devices.
Our results demonstrate that operators are indeed capable of discriminating objects of different stiffness, relying on haptic feedback alone and without any visual feedback. Additionally, our findings suggest that the displacement feedback component 
may enhance discrimination with objects of similar stiffness.
\end{abstract}

\begin{IEEEkeywords}
Teleoperation, Telemanipulation, Robotic Manipulation, Tactile Sensing, Stiffness Perception
\end{IEEEkeywords}

\IEEEpeerreviewmaketitle
\section{Introduction}
\IEEEPARstart{R}{obotic} teleoperation is a fundamental technology for various real-world applications, ranging from healthcare to harsh environments \cite{deng2021review, zhang2022teleoperation, nahri2022review, vitanov2021suite}. 
 A crucial task in teleoperation is telemanipulation, involving manipulation of objects by the teleoperated robot \cite{darvish2023teleoperation}.
Perceiving the physical properties of remote objects is necessary for effective manipulation and information gathering, such as object recognition or assessing specific characteristics (e.g., determining if a manipulated tissue contains a tumor or if a strawberry is ripe).
During bilateral telemanipulation, the human user uses a device (leader) to control a remote robot (follower) and receives feedback from sensors placed on the remote robot or in the environment.

Vision is the dominant modality for sensing remote objects and environments, and visual feedback is easily provided via monitors or Virtual Reality headsets. However, tactile and force sensing are useful complementary modalities and sometimes better than vision for detecting specific physical properties like stiffness. Haptic feedback devices that can effectively render these modalities are still uncommon \cite{Patel10}.

Notably, while there is a large literature about the autonomous robotic perception of object stiffness using tactile sensing \cite{li20tacInfoReview,bednarek21objStifEst,huang21objStifRec}, which can facilitate tactile object recognition \cite{liu17tactileObjRec,kirby22tacObjRec,funabashi24tacObjRec} and tactile assessment of specific object characteristics \cite{ribeiro20tacFruitAssess}, to the best of our knowledge, no studies have explored how human users wearing hand exoskeletons can perceive the stiffness of remote objects during robotic telemanipulation relying solely on remote tactile sensing (follower side) and kinesthetic haptic feedback (leader side).

\begin{figure}[t]
\centering
\includegraphics[width=1.0\linewidth,keepaspectratio]
{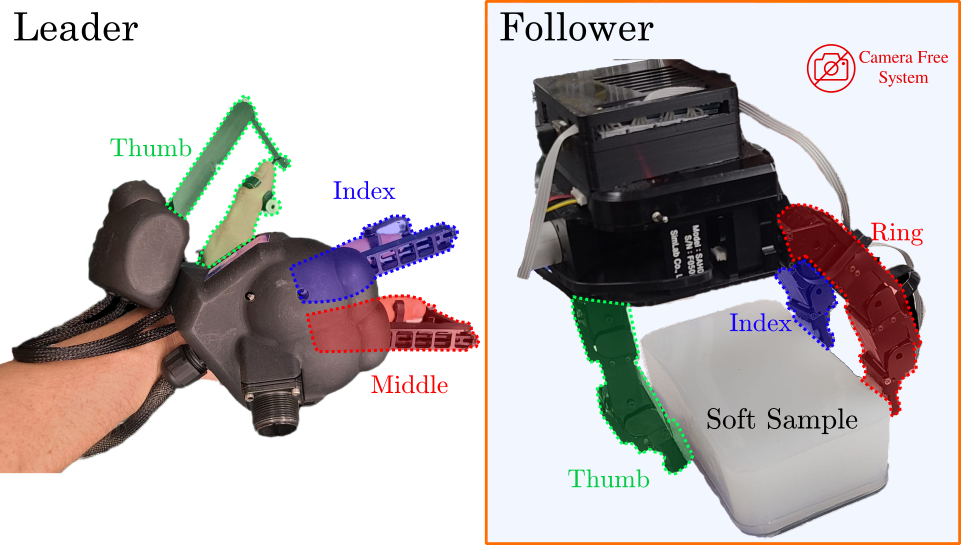}
\caption{The teleoperation setup connecting the Leader HGlove exoskeleton glove to the Allegro robotic hand. Forces measured from the robotic thumb, index, and ring fingers are rendered on the glove's thumb, index, and middle fingers, respectively.
}
\label{hand_mapping}
\end{figure}

To address this gap, in this work we test ten human participants in the task of differentiating remote objects based on their stiffness, by using an advanced bilateral telemanipulation setup with kinesthetic haptic feedback.
The setup is composed of an exoskeletal glove (HGlove \cite{perret2017hglove}) on the leader side, which captures human fingers motion to control the remote robot and provides haptic feedback by applying kinesthetic forces on the human fingers, and a dexterous robotic hand (Allegro Hand) equipped with tactile sensitive fingertips on the follower side, as shown in Fig. \ref{hand_mapping}.
Participants are tasked with squeezing five sample objects differing only in their stiffness (see Section \ref{Sample Description}) using the telemanipulation setup, and determining which is most similar to a given reference (Task ABX) and which of them is softer (Task S), as detailed in Section \ref{Tasks}.
We compare two methods for haptic feedback rendering:
the first method (see Section \ref{sec:method1}) is based solely on the contact forces measured by the tactile sensors, and it was preliminary introduced in \cite{giudici2023feeling}; the second method (newly proposed here, see Section \ref{sec:method2}) considers also the squeezing displacement differences between the human and robot fingers, due to the kinematic mismatch between the leader device and the follower device. 

The main contributions of this work are:
\begin{itemize}

    \item \textbf{Demonstration of Stiffness Perception during Telemanipulation:} We demonstrate that, using a bilateral telemanipulation setup with haptic feedback on the leader and tactile sensing on the follower, ten naive users can discriminate five soft objects differing solely in stiffness by remotely squeezing them with a dexterous robotic hand and perceiving only haptic feedback (no visual cues), as shown in Fig. \ref{fig:BOX_plot_overall}.

    \item \textbf{Introduction of a novel Haptic Feedback method to perceive object stiffness:} 
    In Section \ref{sec:method2}, we introduce a novel method that incorporates finger displacement feedback, in addition to force feedback, to address differences in the squeezing motion between the human operator and the robot caused by kinematic mismatches between the leader and follower devices—a common scenario in telemanipulation.

    \item \textbf{Insights from Statistical Analysis on User Preferences:} Our analysis reveals that incorporating finger displacement feedback benefits naive users primarily in specific circumstances, namely, during more challenging tasks. While the advantage is not consistently significant across all conditions (see Tab. \ref{table:success_rates}), displacement feedback does enhance discrimination performance for objects with similar stiffness (as discussed in  \ref{sub: Sample Pairs Results}).
\end{itemize}

The paper is organized as follows: Section \ref{sec:2_RelWork} reviews related work. In Section \ref{sec:3_Meth}, we describe the robotic setup used in this study, including the haptic stiffness feedback methods. Section \ref{sec:4_Exp} outlines the experimental setup and procedure. The results are analyzed in Section \ref{sec:5_DataAnaly}. In Section \ref{sec:6_Discussion}, we discuss the findings, and Section \ref{sec:7_Conclusion} concludes the paper and suggests future research directions.

\vspace{-0.1cm}
\section{Related Works}
\label{sec:2_RelWork}
A wide range of studies in the literature focuses on automatic stiffness estimation or classfication using tactile sensors \cite{li20tacInfoReview, mavrakis2020estimation, bednarek21objStifEst, omarali2022tactile,huang21objStifRec}, as well as stiffness rendering through various visual technologies for distinguishing soft objects \cite{deng2021review, yamamoto2009tissue, meccariello2016experimental}. However, due to technological and safety limitations, fewer studies have investigated stiffness rendering exclusively through haptic devices.
In this study, we aim to exploit the potential of information obtainable from haptic sensors in a robotic setup, i.e. force and position, to identify a robust strategy to enable the operator to differentiate soft objects during real-world teleoperation experiments without any visual feedback.

\subsection{Teleoperation}
Telemanipulation is a wide field and several comprehensive reviews should be considered, such as \cite{deng2021review, darvish2023teleoperation, si2021review,suomalainen2022survey}. 
A key example of recent advancements in this area is the ANA Avatar XPRIZE, a major non-medical teleoperation competition, that challenged teams to develop systems excelling in intuitiveness, immersiveness, social interaction, robustness, and manipulation capacity across various tasks. 
Notable contributions included a full-body teleoperation platform with mass-spring-based haptic feedback \cite{ParkXprize}, the AvaTRINA Nursebot combining haptic feedback with augmented reality \cite{marques2023commodityXPrize}, and the winning NimbRo Avatar system, which demonstrated robust telepresence and manipulation capabilities \cite{schwarz2023robust}.
Among these, the work \cite{park2024intuitive} introduced an exoskeletal glove for kinesthetic feedback, however, the system provided limited force feedback during the competition, focusing primarily on grasp detection rather than stiffness differentiation. While these contributions advanced telemanipulation technology, none addressed the challenge of stiffness perception or a comparative evaluation of feedback methods. 

\subsection{Stiffness Perception} 
To explore teleoperation systems that render stiffness information for soft objects, the field of telesurgery is particularly relevant. The study \cite{danion2012role} demonstrated the value of haptic feedback in manipulating non-rigid objects. Recent advancements in surgical robotics have integrated haptic feedback to enhance real-time force reflection during procedures \cite{Patel10}. 
While systems like the da Vinci robot have lacked direct haptic feedback, studies using the da Vinci Research Kit (dVRK) have explored solutions \cite{d2021accelerating}, and the upcoming da Vinci 5 model will incorporate certified haptic feedback. The Senhance Surgical System also highlights the benefits of haptic feedback for precise control \cite{Patel10}, while virtual simulations, such as those developed in \cite{KURODA2005216}, have provided realistic force feedback for training purposes. Additionally, haptic techniques using contact force control and virtual springs have been explored to improve transparency and stability \cite{park2006haptic}, while other works \cite{velanas2010human} have focused on adaptive impedance control. Another work \cite{torabi2019application} explored the use of redundant haptic interfaces in teleoperated surgical systems to enhance soft-tissue stiffness discrimination by reducing apparent inertia and improving manipulability, leading to more accurate tissue stiffness perception during virtual palpation tasks.
The review \cite{nagy2019recent} have also addressed soft tissue contact identification with a focus on visual stiffness map obtained by palpation. 
Additionally, some studies have integrated medical instruments with industrial teleoperation systems.
For example, one study \cite{psomopoulou2020evaluation} used a teleoperation system with force feedback to evaluate users' ability to differentiate materials of varying stiffness during a palpation task, utilizing a Haption Virtuose 6D haptic device as the leader and an industrial robot with a da Vinci Endowrist instrument as the follower.
However, our study proposes to use an exoskeleton hand to assess the perception of haptic feedback directly on the fingers, with a larger number of samples and eliminating visual feedback. Furthermore, we will compare two feedback methodologies to assess the role of measured force and compression differences between the leader and follower device.

\section{Methodology}
\label{sec:3_Meth}

\subsection{Teleoperation Setup}
In this study, we utilize a portion of the robotic setup previously described in \cite{giudici2023feeling, mao2024dexskills}. Specifically, our focus on rendering feedback forces on the fingers leads us to limit the teleoperation system to the Allegro robotic hand and the Leader HGlove exoskeleton glove as shown in Fig. \ref{hand_mapping}.
As described in \cite{giudici2023feeling}, each robotic finger of the Follower robotic hand is equipped with a fingertip composed of four magnetic sensors that provide force measurements in three directions: \( F_{x} \), \( F_{y} \), and \( F_{z} \).
These force sensitive fingertips are described in more details here \cite{bonzini2022leveraging}; the sensors are inspired to the design that was initially reported here \cite{tomo2015development} and then further developed \cite{paulino2017low, tomo2017covering}.
In this work, we will use only three fingers of the Allegro Hand robotic hand \( j \in \{\text{thumb, index, ring}\} \), to ensure a one-to-one correspondence with the fingers of the Leader HGlove exoskeletal glove \( i \in \{\text{thumb, index, middle}\} \). This decision does not affect the teleoperation setup's functionality but prevents potential errors due to interaction with the object at multiple closely spaced points. The measured forces will follow the sensor-actuator haptic mapping: thumb to thumb, index to index, and ring to middle.

The maximum force in the \( z \)-direction among the four sensors on the follower's side (subscript  F) is defined as \( F_{\text{F},j} \):

\begin{equation}
\label{eq:max_force}
F_{\text{F},j} = \max(F_{z_1,j}, F_{z_2,j}, F_{z_3,j}, F_{z_4,j})
\end{equation}

The resulting force of each fingertip \( F_{\text{F},j} \) can vary within a range of values greater than 0. However, for our purposes, we define a specific operational range for the force measurements using a minimum threshold \( F_{\text{F}}^{\text{min}} = 30 \) and a maximum threshold \( F_{\text{F}}^{\text{max}} = 1000 \). Thus, the valid range for \( F_{\text{F},j} \) is:

\begin{equation}
F_{\text{F}}^{\text{min}} \leq F_{\text{F},j} \leq F_{\text{F}}^{\text{max}}
\end{equation}

By incorporating these thresholds, we ensure that only meaningful force measurements within the specified range contribute to the feedback mechanisms, thereby enhancing the precision and reliability of the robotic finger's responses. 

\subsection{Haptic Stiffness Feedback Methods}
In this work, we present two methods for effective haptic feedback rendering with the aim of evaluating and characterizing the haptic experience when interacting with soft samples. For this purpose, we will utilize the tactile information from the follower side, \( F_{\text{F},j} \), to generate two different haptic feedback methods on the leader side (subscript L), denoted as \( F_{L,j}^{1} \) and \( F_{L,j}^{2} \).
It is important to highlight that if the sensors do not measure a contact force \( F_{\text{F},j} \) above the minimum threshold, both methods will provide null haptic feedback.
 Thus:

\begin{equation}
\begin{cases} 
F_{L,j}^{1} = 0 \\
F_{L,j}^{2} = 0 
\end{cases} \quad \text{if} \quad F_{\text{F},j} < F_{\text{F}}^{\text{min}}
\end{equation}

\subsubsection{Method I}
\label{sec:method1}
In the proposed Method I, the leader's feedback force \( F_{L,j}^{1} \) is rendered as:

\begin{equation}
F_{L,j}^{1} = \alpha \cdot F_{\text{F},j}
\end{equation}

The constant scalar factor \(\alpha\) is defined as:

\begin{equation}
\alpha = \frac{F_{L,\text{max}}}{F_{\text{F}}^{\text{max}}}
\end{equation}

where \( F_{L,\text{max}} \) represents the maximum force that the leader can apply, set to 5 N. This specification ensures that the maximum rendered force remains within the operational limits of the haptic glove, maintaining system integrity and functionality.

\subsubsection{ Method II}
\label{sec:method2}
In the proposed Method II, the definition of the force rendered by the haptic glove \( F_{L,j}^{2} \) is based on the real-time stiffness estimation through the tactile sensors. The stiffness of the measured samples can be defined as follows:

\begin{equation}
K_{F,j} = \frac{F_{\text{F},j}}{\Delta Z_{F,j}}
\end{equation}

where \( \Delta Z_{F,j} \) represents the displacement in the z-direction of the follower in response to a z-displacement \( \Delta Z_{L,j} \) of each leader's finger during the force measurement.

To relate the displacement of the robotic glove to the displacement of the robotic hand during contact, we define the parameter \( \Delta_Z \) as:

\begin{equation}
\Delta_Z = \frac{\Delta Z_{L,j}}{\Delta Z_{F,j}}    
\end{equation}

Additionally, we introduce the parameter \( \beta \), defined as:

\begin{equation}
\beta = \frac{\Delta Z_{F,j}^{\text{max}}}{\Delta Z_{L,j}^{\text{max}}}
\end{equation}

This parameter allows us to normalize the differences in movement between devices with different kinematics motion ranges.

Consequently, the leader's feedback force \( F_{L,j}^{2} \) is given by:

\begin{equation}
    F_{L,j}^{2} = \alpha \cdot F_{\text{F},j} \cdot \Delta_Z \cdot \beta = F_{L,j}^{1} \cdot \Delta_Z \cdot \beta
\end{equation}

\section{Experiments} 
\label{sec:4_Exp}

\vspace{-0.1cm}
\subsection{Object Description}
\label{Sample Description}
\begin{figure}[t]
\centering
\includegraphics[width=1.0\linewidth,keepaspectratio]
{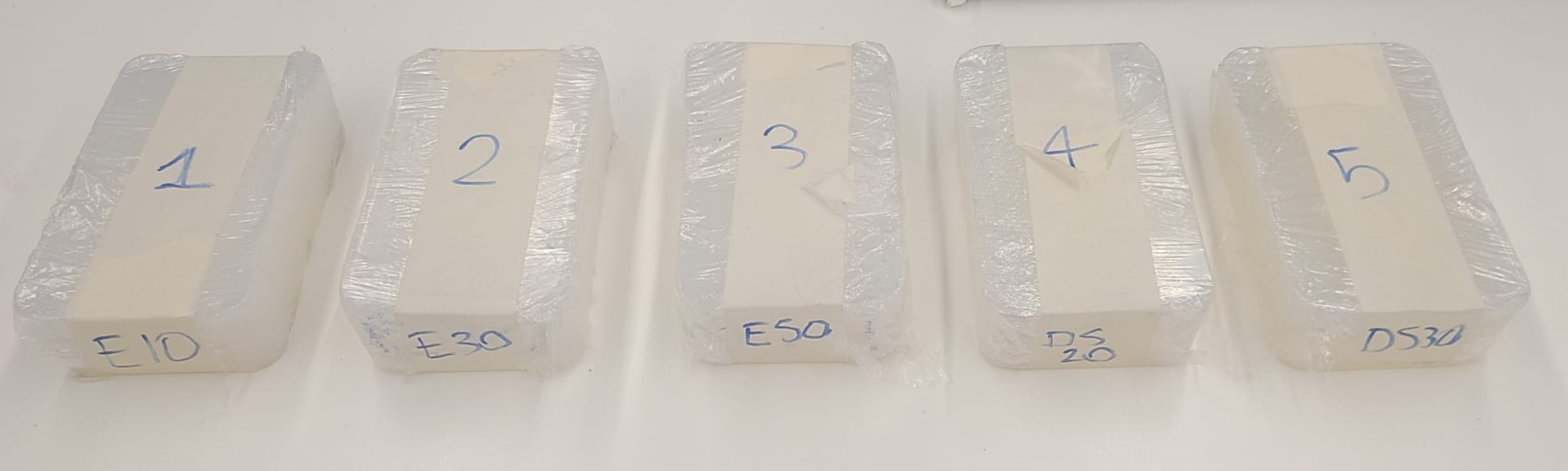}
\caption{Samples of same-sized objects composed of soft materials such as gel and silicone. The samples are numbered from 1 to 5 as their stiffness increases.}
\label{objects}
\end{figure}

In this study, each participant telemanipulated five soft samples of different stiffnesses. These samples were categorized based on their Shore Hardness (SH), a scale that measures the hardness of materials. Specifically, the Shore 00 Hardness Scale is used for very soft materials like rubbers and gels, while the Shore A Hardness Scale measures the hardness of flexible mold rubbers, which can range from soft and flexible to hard and nearly inflexible \cite{qi2003durometer,SmoothOnWebsite}.
The samples were labeled according to their stiffness levels as follows: 
\begin{itemize}
    \item 1-US: Ultra-Soft (Ecoflex 00-10, SH: 00-10)
    \item 2-S: Soft (Ecoflex 00-30, SH: 00-30)
    \item 3-M: Medium (Ecoflex 00-50, SH: 00-50)
    \item 4-LH: Light-Hard (Dragon Skin 20, SH: A-20)
    \item  5-H: Hard (Dragon Skin 30, SH: A-30) .
\end{itemize}
As shown in Fig. \ref{objects}, these labels reflect the increasing stiffness of the samples, providing a standardized measure for comparing the haptic feedback experienced by the participants.

\subsection{Participants Preparation}
In our study on haptic stiffness, the experimental procedure began with adjusting the chair to ensure the participant's comfort and accessibility. 
The study was conducted with ten naive participants.
Participants were selected from a sample of adults 20-35 years old, with no prior physical problems and right-handed. 
Participants were asked to fill out a form with their names and the feedback method used during the experiment as well as to sign a form for approval of the conditions and ethics of the experiments. Subsequently, participants wore the HGlove, with instructions provided on how to proceed if the glove became dislodged during the experiment, which would necessitate repeating the trial. A critical aspect of the experiment was emphasized: participants were instructed to move their fingers in a controlled and slow manner to prevent delays and the occurrence of reflex forces.
\subsection{Training Session}
A short training session, lasting five minutes, was conducted where participants teleoperated the Allegro robotic hand to squeeze an object (sample 3-M) for 90 seconds while visually observing the interaction. Following this, participants squeezed three different samples in a known predefined sequence (sample 1-US, sample 3-M, sample 5-H) for 30 seconds each. Subsequently, a panel was used to hide the scene, preventing participants from seeing the robot and objects.
%
\subsection{Tasks}
\label{Tasks}
The first task is of type \textbf{ABX} \cite{munson1950standardizing}. During this task, participants were presented with two objects ($A$ and $B$) for 10 seconds each, followed by a third object ($X$). They were then asked to determine whether $X$ was more similar to $A$ or $B$. We refer to this task as \textbf{Task ABX}. 

The second task was for participants to answer the following question, "Which object between $A$ and $B$ is softer?". We refer to this task as \textbf{Task S}.

To familiarize with the tasks, two practice examples whose results have not been recorded were given to each participant at the beginning of the session, with the following sample sequences: $[1-US, 3-M, 1-US]$ and $[5-H, 3-M, 3-M]$.

\subsection{Experimental Design}

The experimental session consisted of 24 \textbf{Task ABX} tests and 24 \textbf{Task S} tests, conducted in three sets of eight tests each. The sequence of experiments presented to each participant for both methods is detailed in Table~\ref{tab:abx_sequence}. As illustrated, sequences [1--8] are identical to sequences [17--24], while in sequences [9--16], the values of $A$ and $B$ were inverted, with $X$ remaining unchanged. The values of $X$ were selected to represent the extreme and medium stiffness levels, specifically $X=1$-US, $X=3$-M, and $X=5$-H.

We designed the trials to evaluate four levels of perceptual distance ($D = 1, 2, 3, 4$), each defined by the difference in stiffness levels between stimuli. Each distance level was represented by two distinct pairs of stimuli, and each distance level appeared six times in the experimental session, ensuring balanced representation across all levels. Consequently, each of the eight pairs was tested three times, providing an equitable distribution of trials among the pairs. The specific pairs for each distance level were:

\begin{itemize}
    \item \textbf{D = 1:} (1-US, 2-S \textbf{$\vert$ X=1}), (3-M, 4-LH \textbf{$\vert$ X=3})
    \item \textbf{D = 2:} (1-US, 3-M \textbf{$\vert$ X=3}), (3-M, 5-H \textbf{$\vert$ X=5})
    \item \textbf{D = 3:} (1-US, 4-LH \textbf{$\vert$ X=1}), (2-S, 5-H \textbf{$\vert$ X=5})
    \item \textbf{D = 4:} (1-US, 5-H \textbf{$\vert$ X=1}), (5-H, 1-US \textbf{$\vert$ X=5})
\end{itemize}

In Table~\ref{tab:abx_sequence}, the distance $D$ and the direction of difference are reported; an upward arrow (↑) indicates that the object not similar to $X$ is harder than $X$, while a downward arrow (↓) indicates that it is softer. Participants were unaware that the sequences across the two days were the same and were informed that the sequences were randomly generated. After each set of eight experiments, participants were given a 60-second break during which the glove was readjusted if necessary.
Stimuli within each trial were presented in a randomized order to mitigate sequence effects and potential biases. The assignment of the $X$ stimulus (the target) to match either stimulus $A$ or $B$ was also randomized across trials.
This approach aligns with the requirements of the ABX test, which stipulate that multiple trials must be performed to make statistically confident assertions about a participant's ability to distinguish between stimuli \cite{greenaway2017abx}.
Additionally, limiting the session to 24 trials adheres to guidelines aimed at preventing participant fatigue, thereby maintaining the sensitivity and reliability of the test results.

Two kinesthetic haptic feedback methods were employed: Method I and Method II. Each participant completed two sessions, alternating between the feedback methods. To mitigate any potential bias related to the learning curve, half of the participants started the first session with Method I, while the other half began with Method II.
Each experimental session lasted approximately 40 to 45 minutes, including about 10 minutes for setup, training, and rest, and approximately 30 minutes dedicated to the experiments.

\section{Data Analysis}
\label{sec:5_DataAnaly}
This section is organized to highlight the different analyses we conducted to investigate the experimental results. In subsection \ref{sub: Confidence level}, we define the reference confidence values of our tests to demonstrate their relevance. In subsection \ref{sub: Overall Performance Comparison}, we analyze and compare the overall performance of the two feedback methods. In subsection \ref{sub:daily Perfomance}, we extend the analysis by separating the data according to the two days of experimentation. In subsection \ref{sub: ANOVA}, we evaluate the effects of various experimental variables on the success rate. Finally, in subsection \ref{sub: Sample Pairs Results}, we examine whether performance differences exist when comparing different sample pairs.

\vspace{-0.1cm}
\subsection{Statistical Confidence Level}
\label{sub: Confidence level}
In the ABX test, each trial has a 50\% chance of a correct response under the null hypothesis of random guessing. Using the binomial distribution with $n = 24$ trials and success probability $p = 0.5$, the probability of obtaining at least 16 correct responses by chance (66.66\% of success rate) is approximately 10.6\%:

\begin{equation}
P(X \geq 16) = \sum_{k=16}^{24} \binom{24}{k} (0.5)^{24} \approx 10.6\%.
\end{equation}

This corresponds to a confidence level of approximately 89.4\%. While this does not meet the threshold for statistical significance at the 90\% confidence level, it suggests a trend toward significance in participants' ability to discriminate between stimuli.

To achieve statistical significance above the 95\% confidence level, participants need at least 17 correct responses on average (70.8\% of success rate), as the probability of obtaining this result by chance is approximately 4.3\%:

\begin{equation}
P(X \geq 17) \approx 4.3\%.
\end{equation}

Therefore, we can reject the null hypothesis at the 95\% confidence level, indicating a statistically significant ability to discriminate between stimuli.

\begin{table*}[h]
\centering
\caption{Sequence of ABX Tests}
\begin{tabular}{| c| c| c| c| c| c| c| c| c|| c| c| c| c| c| c| c| c|| c| c| c| c| c| c| c| c|}
\hline
\textbf{Test} & \textbf{1} & \textbf{2} & \textbf{3} & \textbf{4} & \textbf{5} & \textbf{6} & \textbf{7} & \textbf{8} & \textbf{9} & \textbf{10} & \textbf{11} & \textbf{12} & \textbf{13} & \textbf{14} & \textbf{15} & \textbf{16} & \textbf{17} & \textbf{18} & \textbf{19} & \textbf{20} & \textbf{21} & \textbf{22} & \textbf{23} & \textbf{24} \\ \hline

\textbf{A} & 1 & 2 & 4 & 3 & 5 & 2 & 1 & 3 & 5 & 5 & 3 & 1 & 1 & 1 & 4 & 5 & 1 & 2 & 4 & 3 & 5 & 2 & 1 & 3 \\ 
\textbf{B} & 5 & 5 & 3 & 1 & 1 & 1 & 4 & 5 & 1 & 2 & 4 & 3 & 5 & 2 & 1 & 3 & 5 & 5 & 3 & 1 & 1 & 1 & 4 & 5 \\ 
\textbf{X} & 1 & 5 & 3 & 3 & 5 & 1 & 1 & 5 & 1 & 5 & 3 & 3 & 5 & 1 & 1 & 5 & 1 & 5 & 3 & 3 & 5 & 1 & 1 & 5 \\ \hline \hline
\textbf{D} & 4 & 3 & 1 & 2 & 4 & 1 & 3 & 2 & 4 & 3 & 1 & 2 & 4 & 1 & 3 & 2 & 4 & 3 & 1 & 2 & 4 & 1 & 3 & 2 \\ 
\textbf{} & ↑ & ↓ & ↑ & ↓ & ↓ & ↑ & ↑ & ↓ & ↑ & ↓ & ↑ & ↓ & ↓ & ↑ & ↑ & ↓ & ↑ & ↓ & ↑ & ↓ & ↓ & ↑ & ↑ & ↓ \\ \hline
\end{tabular}
\label{tab:abx_sequence}
\end{table*}


\begin{table}[htbp]
\centering
\caption{Mean Success Rates, Variances, and Standard Deviations for Tasks X and S by Method}
\begin{tabular}{|c|c|c|c|c|}
\hline
\textbf{Task} & \textbf{Method} & \parbox[c]{2cm}{\centering \hspace{0.1cm} \textbf{Mean Success Rate (\%)} \hspace{0.1cm}} & \textbf{Variance} & \textbf{Std Deviation} \\ \hline
\textbf{ABX} & \textbf{1}  & \textbf{74.16} & \textbf{64.83} & \textbf{8.05} \\ \hline
\textbf{ABX} & \textbf{2}  & \textbf{74.58} & \textbf{21.04} & \textbf{4.59} \\ \hline
\textbf{S} & \textbf{1}  & \textbf{67.91} & \textbf{81.20} & \textbf{9.01} \\ \hline
\textbf{S} & \textbf{2}  & \textbf{64.58} & \textbf{159.14} & \textbf{12.62} \\ \hline
\end{tabular}
\\[0.2cm] 
\label{table:success_rates}
\end{table}

\subsection{Overall Performance Comparison}
\label{sub: Overall Performance Comparison}
\begin{figure}[h]
    \centering
    \includegraphics[width=0.9\linewidth]{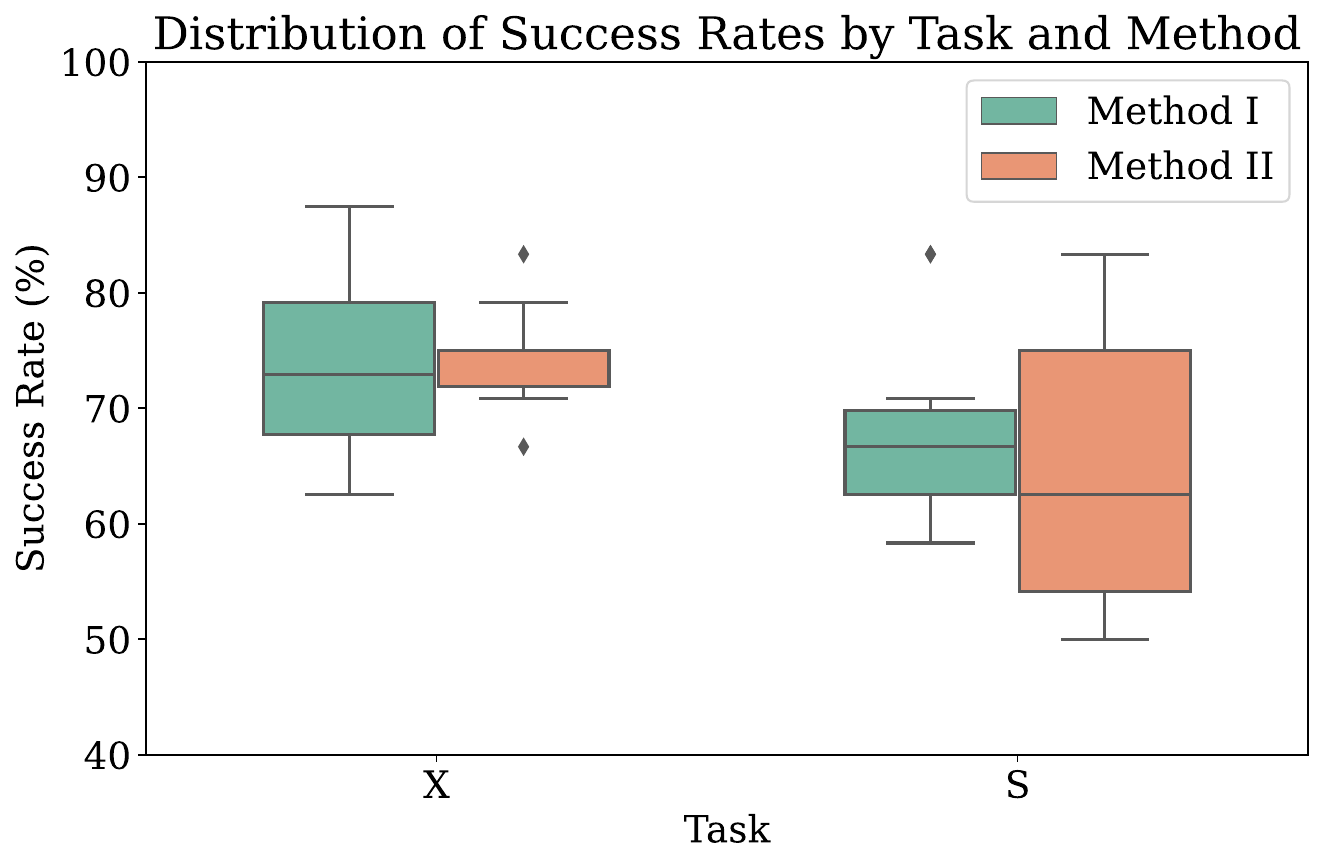}
    \caption{The graph compares the overall success rate for each task by method, with Task ABX results on the left and Task S results on the right, without differentiating between experimental days. $\blacklozenge$ are the outliers.}
    \label{fig:BOX_plot_overall}
\end{figure}

The box plot in Fig. \ref{fig:BOX_plot_overall} depicts the overall performance of the participants for \textbf{Task ABX} and \textbf{Task S} using the respective methods, regardless of the day of the experimental session.

\begin{itemize}
    \item \textbf{Task ABX Performance:} Both Method I and Method II exhibit relatively high success rates for \textbf{Task ABX}, with mean success rates around 75\%. However, Method II shows lower variability, indicating more consistent performance for participants, as evidenced by lower variance and standard deviation.

    \item \textbf{Task S Performance:} Similarly, for \textbf{Task S}, the success rates of both methods are close. Method I has a slightly higher median success rate (67.91\%) compared to Method II (64.58\%). However, Method II exhibits a broader spread in performance, indicating greater variability among participants. The presence of potential outliers suggests that some participants faced difficulties with Method II in \textbf{Task S}.
  
\end{itemize}

The box plot indicates that for \textbf{Task ABX}, Method II is more consistent compared to Method I, suggesting that participants may find it easier to maintain stable performance with Method II. For \textbf{Task S}, while Method I has a slightly higher median, Method II shows much higher variability, which could suggest that individual differences or specific challenges inherent to \textbf{Task S} are more pronounced with Method II.

\vspace{-0.2cm}
\subsection{Daily Performance Comparison}
\label{sub:daily Perfomance}
One limitation of the previous analysis is that it does not account for potential learning effects between Day 1 and Day 2. To address this, we separated the participants into two groups:
\begin{itemize}
    \item \textbf{Group 1}: Participants who started with Method I (blue) on Day 1 and continued with Method II(red) on Day 2.
    \item \textbf{Group 2}: Participants who started with Method II (red) on Day 1 and continued with Method I (blue) on Day 2.
\end{itemize}

\begin{figure}[h]
    \centering
    \includegraphics[width=1\linewidth]{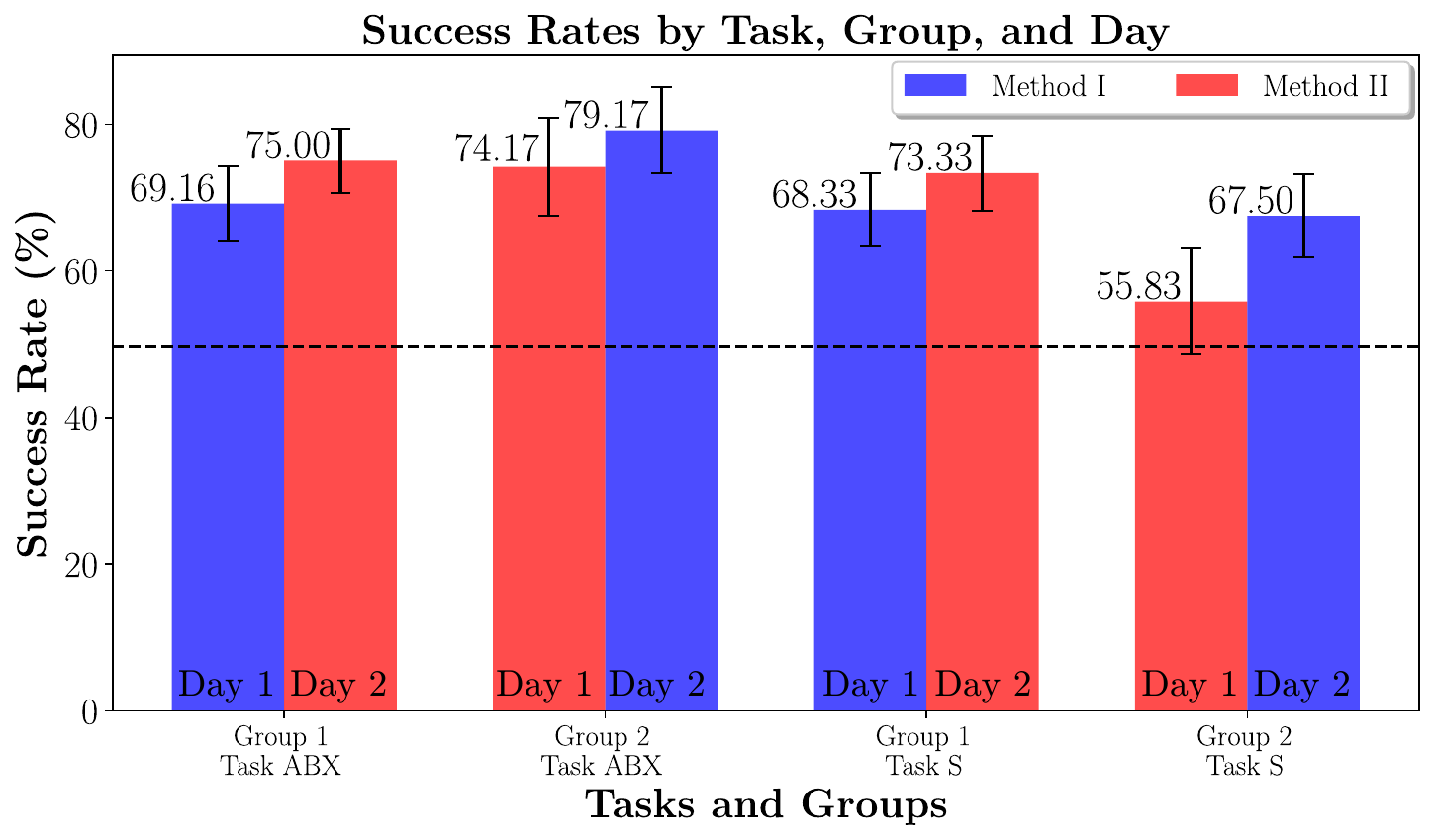}
    \caption{Task success rates for both groups, subdivided by experimental days and methods applied. }
    \label{fig:GroupDayTask}
\end{figure}

 The bar plot in Fig. \ref{fig:GroupDayTask} illustrates the success rates of the two groups across the two tasks (\textbf{Task ABX} and \textbf{Task S}) with distinct methods applied on different days. Each bar represents the mean success rate for a specific group on a specific day and task.

The graph reveals several key observations:
\begin{itemize}
    \item Both groups show an average improvement in performance between the first and second day of testing for each task. Group 1 improved in \textbf{Task ABX} from 69.16\% to 75\%, an increase of approximately 6.0\%, and in \textbf{Task S} from 68.33\% to 73.33\%. Group 2 also shows an improvement from 74.17\% to 79.17\% between the first and second day on \textbf{Task ABX} and a noticeable improvement in \textbf{Task S} from 55.83\% to 67.50\%.

    \item  As observed, Group 1 exhibited similar performance across both tasks. In contrast, Group 2 outperformed Group 1 in \textbf{Task ABX} but showed generally lower performance in \textbf{Task S}.
    
\end{itemize}

In conclusion, there is a noticeable improvement in understanding the tasks between Day 1 and Day 2, especially for \textbf{Task S} (which is generally more difficult).

\vspace{-0.1cm}
\subsection{Analysis of Variance (ANOVA)}
\label{sub: ANOVA}
A two-way ANOVA—a statistical method that evaluates the main effects of two independent variables and their interaction on a dependent variable—was conducted to examine the effects of \textit{Group}, \textit{Day}, and \textit{Task} on the mean \textit{Success Rate}, along with their interaction effects. The analysis tested the following main effects and interactions:

\begin{itemize}
    \item \textbf{Group}: Evaluating the influence of participant group on success rates.
    \item \textbf{Day}: Assessing the impact of the day on which the task was performed on success rates.
    \item \textbf{Task}: Determining the effect of the type of task (X or S) on success rates.
    \item \textbf{Interactions}: Investigating interactions between \textit{Group} and \textit{Day}, \textit{Group} and \textit{Task}, and \textit{Day} and \textit{Task}.
\end{itemize}

The results of the ANOVA are presented in Table \ref{Tab:Anova3} and the results are summarized as follows:

\begin{table}[h]
\centering
\caption{ANOVA Results for Success Rates by Group, Day, and Task}
\label{tab:anova_results}
\begin{tabular}{lcccc}
\hline
\textbf{Source}                & \textbf{Sum of Squares} & \textbf{df} & \textbf{F-Statistic} & \textbf{p-value} \\
\hline
\textit{Group}                 & 10.47                   & 1           & 1.48                 & 0.437            \\
\textit{Day}                   & 94.60                   & 1           & 13.42                & 0.170            \\
\textit{Task}                  & 132.11                  & 1           & 18.74                & 0.145            \\
\textit{Group:Day}             & 4.25                    & 1           & 0.60                 & 0.580            \\
\textit{Group:Task}            & 94.60                   & 1           & 13.42                & 0.170            \\
\textit{Day:Task}              & 4.25                    & 1           & 0.60                 & 0.580            \\
\textit{Residual}              & 7.05                    & 1           & -                    & -                \\
\hline
\label{Tab:Anova3}
\end{tabular}
\end{table}

\begin{itemize}
    \item The effect of \textbf{Group} on the \textit{Success Rate} was not statistically significant, \( F = 1.48 \), \( p = 0.437 \), indicating that group assignment did not significantly influence participants' success rates.
    \item The effect of \textbf{Day} on the \textit{Success Rate} was also not statistically significant, \( F = 13.42 \), \( p = 0.170 \), suggesting that the day of task performance did not significantly affect success rates.
    \item The effect of \textbf{Task} on the \textit{Success Rate} was not statistically significant, \( F = 18.74 \), \( p = 0.145 \), implying that success rates did not differ significantly between Task X and Task S.
    \item The interactions between \textbf{Group} and \textbf{Day} (\( F = 0.60 \), \( p = 0.580 \)), \textbf{Group} and \textbf{Task} (\( F = 13.42 \), \( p = 0.170 \)), and \textbf{Day} and \textbf{Task} (\( F = 0.60 \), \( p = 0.580 \)) were all not statistically significant.

\end{itemize}

The ANOVA results indicate that there were no significant differences in success rates across different groups, days, or tasks. Furthermore, the lack of statistically significant interaction effects suggests that the combination of these factors did not meaningfully influence the outcomes. This implies that variations in success rates cannot be attributed to the specific grouping of participants or the task types.
The relatively high F-values observed for some factors (such as \textit{Day} and \textit{Task}) paired with non-significant p-values may suggest that, although some differences were present, they were not substantial enough to achieve statistical significance.
The analysis yielded a residual sum of squares of 7.05 indicating that a substantial portion of the variability in the success rates remains unexplained by the factors of Group, Day, Task, or their interactions.

\subsection{Performance and Statistical Analysis Across Sample Pairs}
\label{sub: Sample Pairs Results}

As presented in the section \ref{Sample Description}, the samples utilised during the experiments are classified into seven different pairs, which are further grouped according to their difference in stiffness (denoted as \textbf{D} as shown in Table \ref{tab:abx_sequence}). 

To assess the suitability for parametric testing, we first evaluated normality and homogeneity of variances in success rates for each object pair independently and method using the Shapiro-Wilk and Levene’s tests, respectively. The Shapiro-Wilk results indicated significant deviations from normality (\( p < 0.05 \)) in most groups, and Levene's test showed inconsistent homogeneity across pairs and tasks, suggesting that parametric assumptions were not met.

\begin{figure}[h]
    \centering
    \includegraphics[width=0.35\textwidth]{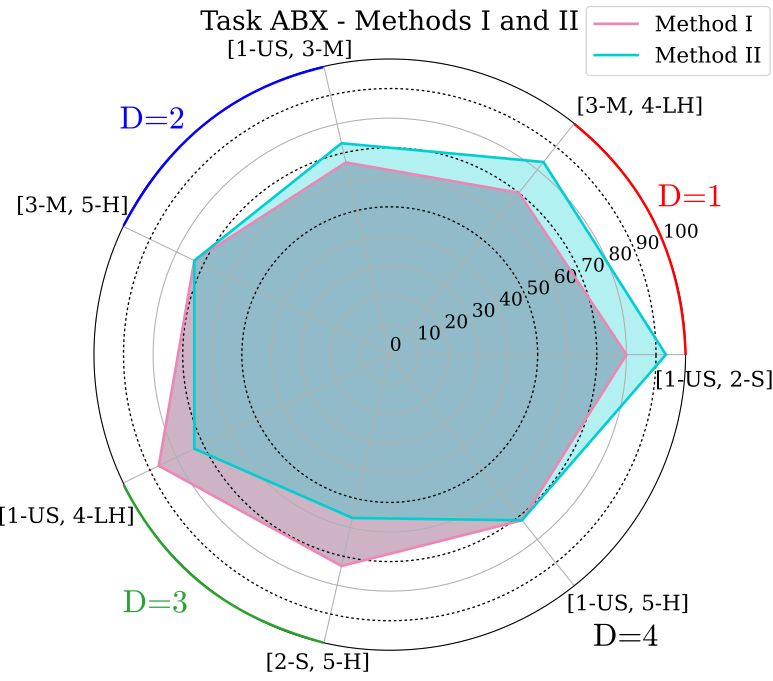}
    \caption{Spider plot showing Success Rate performance for Method I and II effectiveness in Task ABX for object pairs.}
    \label{fig:spider_plot_X}
\end{figure}

\begin{figure}[h]
    \centering
    \includegraphics[width=0.35\textwidth]{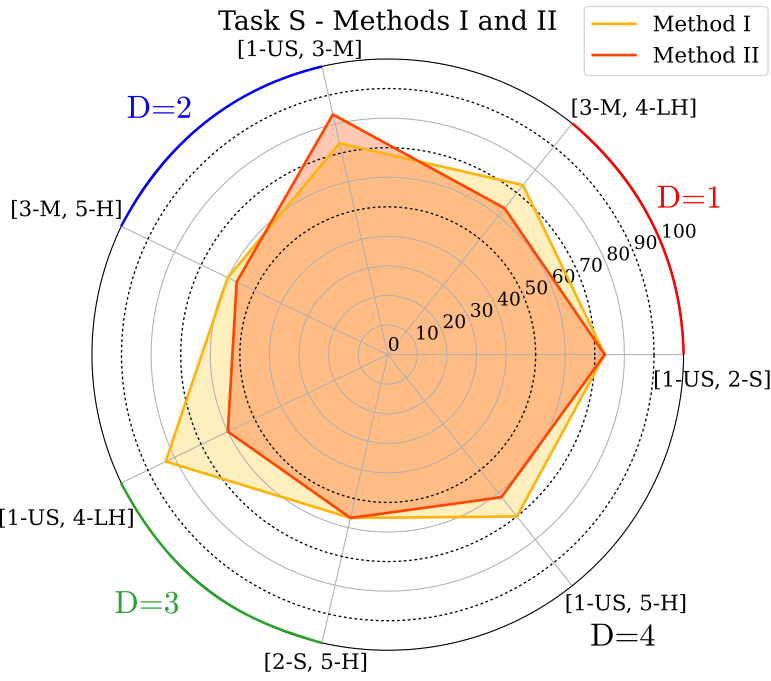}
    \caption{Spider plot showing Success Rate performance for Method I and II effectiveness in Task S for object pairs.}
    \label{fig:spider_plot_S}
\end{figure}

Given these results, we employed the Mann-Whitney U test (MWU), a non-parametric alternative to the t-test, to compare success rates between Method I and Method II for each object pair and task (ABX and S). This test ranks observations across both methods and calculates the \textbf{U} statistic, indicating the extent of difference in rank distributions between the methods. With statistical significance set at \( p < 0.05 \), only pair (1-US, 4-LH) in \textbf{Task S} showed a significant difference between the methods (\( p = 0.048 \)), suggesting differing outcomes for this pair. For all other pairs, the analysis showed \( p > 0.05 \), indicating comparable success rates between the methods across pairs and tasks. To visualize all these findings, we generated spider plots, shown in Fig. \ref{fig:spider_plot_X} and Fig. \ref{fig:spider_plot_S}, which depict success rates across object pairs for each method. 
\vspace{-0.2cm}
\vspace{-0.45cm}

To further explore these differences, we aggregated object pairs by stiffness distance into three groups: \textbf{D = 1}, \textbf{D = 2}, and \textbf{D = 3}. The MWU test was applied to each distance group independently for both ABX and S tasks. Although no group reached \( p < 0.05 \) significance, results for the ABX task produced p-values close to this threshold, \textbf{D = 1} (\( p = 0.061 \)) and \textbf{D = 3} (\( p = 0.067 \)), suggesting a potential trend toward significance. In light of these findings, we computed mean success rates for Method I and Method II within each distance group to better understand performance patterns.  For the ABX task, Method II showed higher mean success rates than Method I for \textbf{ D = 1 } (88\% vs. 75\%), while Method I outperformed Method II for \textbf{ D = 3 } (80\% vs. 65\%). 
In the S task, Method I consistently demonstrated higher mean success rates for both \textbf{ D = 1 } (73\% vs. 68\%) and \textbf{ D = 3 } (65\% vs. 58\%)

These results suggest that in \textbf{Task ABX} Method II may be better suited for shorter distances (\textbf{D = 1}) and slightly better for (\textbf{D = 2}) , while Method I tends to yield better success rates at greater distances (\textbf{D = 3}) across both tasks. The spider plots in Fig. \ref{fig:spider_plot_X} and Fig. \ref{fig:spider_plot_S} visually reinforce these observations, confirming the statistical trends identified through the MWU test.
In \textbf{Task S} the distinction between Methods I and II is less pronounced, with generally lower success rates than in \textbf{Task ABX}. In line with statistical findings, Method I demonstrated slightly higher success rates in pairs involving the softest object (1-US) paired with the hardest objects (4-LH and 5-H), although no statistically significant differences were confirmed beyond the single significant pair (1-US, 4-LH).

\section{Discussion}
\label{sec:6_Discussion}
In this study, we proposed and evaluated two methodologies for effectively rendering haptic stiffness feedback based on tactile contact information from sensors positioned at the fingertips of a robotic hand teleoperated by an exoskeleton glove. The first methodology, referred to as Method I, utilized haptic feedback proportional solely to the contact forces, while the second methodology, referred to as Method II, introduced an additional factor that considered the displacement of the fingers as the object was compressed, thereby compensating for the kinematic mismatch between the exoskeleton and the robotic hand.
The study demonstrated that by employing the presented telerobotic setup it is possible to achieve a significant level of success in discriminating between two soft objects using only contact force information, with a success rate of approximately 75\%, therefore with a confidence higher than 95\%, for both methodologies in \textbf{Task ABX}. Furthermore, in \textbf{Task S}, where participants were asked to discriminate which object was softer, the study showed an average success rate of around 65\% across both methods, therefore with a confidence close to 90\%.
Overall, all participants performed above the 50\% chance level. In addition, it should be noted that there was a noticeable outlier, excluded from the statistical analysis, who achieved a remarkable
95\% success rate in \textbf{Task ABX} and approximately 80\% in
\textbf{Task S}.

A notable finding was that incorporating the displacement factor in Method II improved the discrimination performance only in the most difficult tasks. While Method II did not yield consistently significant improvements over the force-proportional feedback of Method I, suggesting that in general the force feedback component is the most important one, Method II shows improved performance with objects of similar stiffness, where displacement feedback seems to enhance discrimination; instead, Method I proved to be especially effective when objects have large stiffness differences.

\section{Conclusion}
\label{sec:7_Conclusion}
In conclusion, the findings from this study demonstrate that it is possible to provide effective haptic feedback of object stiffness, leading to a significant success rate in object discrimination, which can potentially be enhanced with intensive training. Moreover, our experience underscores how the kinematic differences between the leader and follower devices affect telemanipulation performances, especially in tasks that are more challenging, and suggest that a feedback method accounting for these differences could mitigate this issue.

\ifCLASSOPTIONcaptionsoff
  \newpage
\fi

\bibliographystyle{IEEEtran}
\bibliography{paper}
\end{document}